\documentclass{article}

\usepackage{PRIMEarxiv}

\usepackage{multirow}
\usepackage{amssymb}
\usepackage{amsmath}
\usepackage{xcolor}
\usepackage{hyperref}

\DeclareMathOperator*{\argmax}{argmax}

\usepackage[utf8]{inputenc} 
\usepackage[T1]{fontenc}    
\usepackage{hyperref}       
\usepackage{url}            
\usepackage{booktabs}       
\usepackage{amsfonts}       
\usepackage{nicefrac}       
\usepackage{microtype}      
\usepackage{lipsum}
\usepackage{fancyhdr}       
\usepackage{graphicx}       

\usepackage{graphicx}
\usepackage{subcaption}
\usepackage{url}
\usepackage{xcolor}
\usepackage{adjustbox}

\usepackage{amssymb}
\usepackage{amsmath}

\title{Continual Learning for Behavior-based Driver Identification}

\author{
  Mattia Fanan \\
  University of Padova \\
  Padova, Italy \\
  \texttt{mattia.fanan@phd.unipd.it} \\
   \And
  Davide Dalle Pezze \\
  University of Padova \\
  Padova, Italy \\
  \texttt{davide.dallepezze@unipd.it} \\
   \And
  Emad Efatinasab \\
  University of Padova \\
  Padova, Italy \\
  \texttt{emad.efatinasab@phd.unipd.it } \\
   \And
  Ruggero Carli \\
  University of Padova \\
  Padova, Italy \\
  \texttt{ruggero.carli@unipd.it} \\
   \And
  Mirco Rampazzo \\
  University of Padova \\
  Padova, Italy \\
  \texttt{mirco.rampazzo@unipd.it} \\
   \And
  Gian Antonio Susto \\
  University of Padova \\
  Padova, Italy \\
  \texttt{gianantonio.susto@unipd.it} \\
}

\begin{document}
\maketitle

\begin{abstract}
Behavior-based Driver Identification is an emerging technology that recognizes drivers based on their unique driving behaviors, offering important applications such as vehicle theft prevention and personalized driving experiences.
However, most studies fail to account for the real-world challenges of deploying Deep Learning models within vehicles. 
These challenges include operating under limited computational resources, adapting to new drivers, and changes in driving behavior over time.
The objective of this study is to evaluate if Continual Learning (CL) is well-suited to address these challenges, as it enables models to retain previously learned knowledge while continually adapting with minimal computational overhead and resource requirements. 
We tested several CL techniques across three scenarios of increasing complexity based on the well-known OCSLab dataset.
This work provides an important step forward in scalable driver identification solutions, demonstrating that CL approaches, such as DER, can obtain strong performance, with only an 11\% reduction in accuracy compared to the static scenario. 
Furthermore, to enhance the performance, we propose two new methods, SmooER and SmooDER, that leverage the temporal continuity of driver identity over time to enhance classification accuracy. 
Our novel method, SmooDER, achieves optimal results with only a 2\% reduction compared to the 11\% of the DER approach. 
In conclusion, this study proves the feasibility of CL approaches to address the challenges of Driver Identification in dynamic environments, making them suitable for deployment on cloud infrastructure or directly within vehicles.

\end{abstract}


\section{Introduction}
Behavior-based driver identification relies on driving patterns from sensors' data to determine who is operating the vehicle. This involves collecting information such as acceleration, steering angles, braking patterns, and other driving behaviors over time, followed by inference to determine the driver's identity. Several researchers have proposed behavior-based authentication methods \cite{kang2019automobile, ravi2022driver, TSENG2023105571}, with many recent works utilizing Deep Learning to improve accuracy and reliability \cite{lin2018driver, girma2019driver,xun2019automobile,el2019improving, gahr2019driver,azadani2020performance, abu2020livedi}.

This approach provides significant advantages. For instance, it can help solve the ongoing issue of \textit{car theft}, which is still a problem despite many technological advances. Traditional vehicle authentication methods, often relying on physical or wireless keys, have shown vulnerabilities \cite{garcia2016lock, francillon2011relay}. 
With such technology, \textit{car rental agencies} and \textit{fleet managers} can ensure that only authorized individuals operate vehicles, enhancing fleet management efficiency. 
Moreover, this approach enables personalized services such as \textit{customized in-car experiences}, \textit{enhanced roadside assistance}, and \textit{optimized maintenance schedules}.

While the literature extensively explores the potential applications of behavior-based driver identification systems, challenges specifically related to real-world implementation are often overlooked. 
Many of these systems rely on Machine Learning (ML) and Deep Learning (DL) models for classification, which are typically trained offline on comprehensive datasets of drivers. This approach presents several issues. For instance, gathering sufficient training data can be time-consuming, leaving vehicles vulnerable to theft during this period \cite{efatinasab2024authentication, thesis}.
In scenarios like car sharing or rental services, a complete retraining policy is impractical, as training the model with all data each time a new driver is added can strain resources such as GPU and memory, especially in resource-constrained environments like vehicles. Even with cloud storage, the time required and costs can be prohibitive. 
Moreover, fine-tuning the model using only new data fails to address the issue, as it can lead to Catastrophic Forgetting (CF), where the model's performance on previously learned tasks deteriorates as it adapts to new information \cite{de2021continual, mccloskey1989catastrophic}.

\begin{figure}[thbp]
  \centering
    \includegraphics[width=\textwidth]{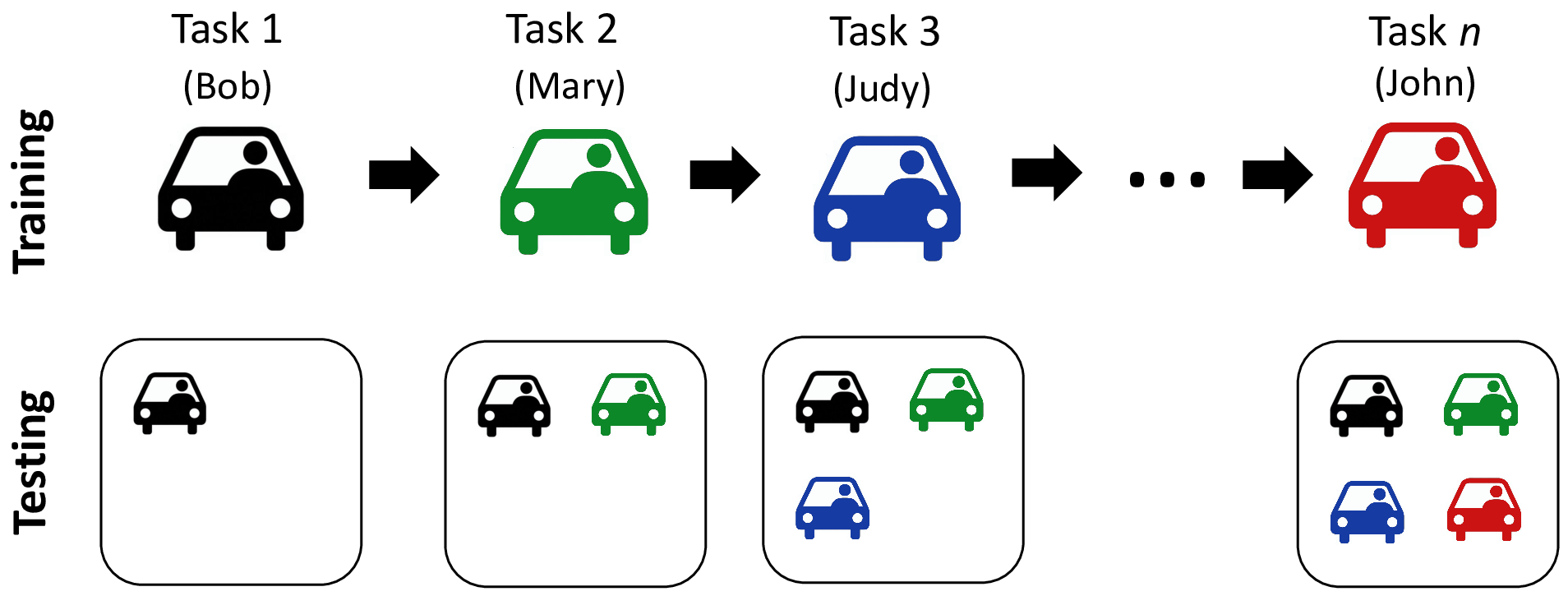}
    \caption{
    General representation of the experimental setup. Since the goal is to learn new tasks as they arrive while retaining knowledge from previous ones, the test set for each task consists of all the test data from drivers whose labels already appeared.}
    \label{fig:CL_driver}
\end{figure}

To address these challenges, we propose applying Continual Learning (CL) techniques to the Driver Identification problem. Being the primary goal of CL to adapt to incoming data while preserving previously acquired knowledge, it is specifically designed to tackle the issue of Catastrophic Forgetting using resources comparable to fine-tuning. 
Therefore, we study scenarios where the models learn incrementally from new drivers (see Fig. \ref{fig:CL_driver} for the general framework).
Specifically, we propose an evaluation across three scenarios of increasing complexity and realism, based on the well-known OCSLab dataset \cite{ocslab}.
The first scenario represents a classical CL setting, while the other two modify how the data is presented to the system, making them increasingly reflective of real-world conditions. Interestingly, augmenting the similarities to real-world scenarios not only introduces challenges but also opportunities. For instance, a driver is not expected to change every few minutes. We leverage this continuity to introduce two new methods SmooER and SmooDER, which aim at further narrowing the performance gap compared to complete retraining.

This work provides an important step forward in scalable driver identification solutions, demonstrating that CL approaches, such as DER, can obtain strong performance, with only an 11\% reduction in accuracy compared to the static scenario.
Furthermore, our novel method, SmooDER, achieves optimal results with only a 2\% reduction compared to the 11\% of the DER approach. 
Such results prove the feasibility of CL approaches to address the challenges of Driver Identification in dynamic environments, making them suitable for deployment on cloud infrastructure or directly within vehicles.
Furthermore, we have made our implementation publicly available \footnote{https://github.com/MattiaFanan/CL-DriverIdentification}, for enhanced reproducibility and to facilitate future research in the field.

\noindent
Our contributions are summarized as follows:
\begin{itemize}
\item We investigate the Driver Identification problem in the CL framework.
\item We propose and implement on the OCSLab dataset three new scenarios that progressively increase in complexity to incorporate additional challenges and make them progressively more similar to a real application of Driver Identification in a physical vehicle.
\item We evaluate the efficacy of CL techniques in the three scenarios, proving the feasibility of their deployment on cloud infrastructure or directly within vehicles.
\item We propose two novel methods, SmooER and SmooDER, that exploit the driver ID continuity expected in deployment, achieving superior results compared to the other CL approaches.
\end{itemize}

The rest of this article is organized as follows:  
Section \ref{sec:related_work} reviews the literature on the Driver Identification problem and the CL framework. 
Section \ref{sec:methodology} describes our proposed scenarios, the CL approaches we tested, and our new methods, SmooDER and SmooER.
Section \ref{sec:experimental_setting} details the experimental setup, including the dataset, training configuration, and model hyperparameters. 
Section \ref{sec:results} presents the results for each scenario. 
Finally, Section \ref{sec:conclusion} discusses the findings and suggests future research directions in this field.

\section{Related Work}
\label{sec:related_work}
We provide in Sec. \ref{subsec:driver_identification} the studies present in the literature concerning the behavior-based driver identification problem.
Subsequently, Sec. \ref{subsec:continual_learning} will provide an overview of the Continual Learning topic, highlighting the main scenarios and methods considered in the field.

\subsection{Driver Identification}
\label{subsec:driver_identification}
A significant amount of research addresses the challenge of identifying the driver based on behavioral data. This issue has various real-world applications, such as reducing vehicle theft and enhancing personalized driving experiences. For example, collecting and analyzing these data can pave the way for optimizing driver-based insurance schemes by dynamically calculating risk factors in real time~\cite{cura2020driver, nyt_behavior}. Moreover, this opens avenues for adjusting vehicle parameters based on user preferences, as demonstrated in previous studies~\cite{el2019improving}. 

Most methodologies gather Controller Area Network (CAN) bus data via the On-Board Diagnostics (OBD-II) port, utilizing its variety of sensors and actuators to assess the vehicle's condition. 
While many researchers do not share the datasets they utilize and gather, most studies use publicly available datasets like OCSLab~\cite{ocslab, kwak2016know}. OCSLab features 54 sensors from the CAN bus and includes data from ten drivers. 
Other datasets, although less commonly used, incorporate different data types, such as Global Positioning System (GPS)~\cite{rahim2019zero} data and stability measurements~\cite{ahmadi2021optimising}. 

Many works use Artificial Intelligence (AI) and ML algorithms for the driver identification problem. For example, \cite{8863987} proposes a method for the identification of drivers using driving dynamics signals available in production cars. Their system collects and filters the sensing data in sliding windows, computes statistical and spectral features, performs driver identification for each window frame through classification, and finally aggregates individual predictions to provide a single prediction for the entire trip. Erzin et al.~\cite{erzin2006multimodal} explore the amalgamation of various features, including pedal pressure, vehicle speed, engine speed, and steering angle, to discern driver identity. 

Various works have employed classical machine learning algorithms, such as Random Forest (RF)~\cite{ahmadi2021optimising, hallac2016driver, ezzini2018behind, rahim2019zero}, Support Vector Machine (SVM)~\cite{hallac2016driver, ezzini2018behind, rahim2019zero, marchegiani2018long, burton2016driver, azadani2021driver}, K-Nearest Neighbors (KNN)~\cite{kwak2016know, lin2018driver, hallac2016driver, ezzini2018behind, rahim2019zero}, and K-means~\cite{kang2019automobile}, for the task of driver identification. 
Marchegiani et al.~\cite{marchegiani2018long} leverage CAN bus data from an electric vehicle to study pedal operation patterns and GPS traces across different drivers on the same route. Their research aims to develop a user authentication system using ML techniques, such as SVM and Universal Background Model (UBM).

Furthermore, there is increasing emphasis in the literature on the use of Recurrent Neural Networks (RNNs)~\cite{el2019improving, gahr2019driver, like_them, girma2019driver}, including Long Short-Term Memory networks (LSTMs)~\cite{azadani2020performance, ravi2022driver, abu2020livedi, el2019improving}. Their widespread adoption underscores their effectiveness in capturing temporal dependencies and modeling sequential data, thereby improving the accuracy and reliability of behavior-based driver identification systems.
Xun et al.~\cite{xun2019automobile} propose driver fingerprinting, utilizing CAN bus data for real-time user authentication. Their method employs a combination of Convolutional Neural Networks (CNN) and Support Vector Domain Description (SVDD) to detect unauthorized drivers. The authors in \cite{chen2019driver} introduce an unsupervised method using a three-layer non-negativity-constrained autoencoder to find the best sliding window size. They then employ a deep autoencoder to extract hidden driving behavior features for better driver identification.

\subsection{Continual Learning}
\label{subsec:continual_learning}
Commonly, a Machine Learning model is built on the premise that the training dataset is representative of the reality one intends to model. However, in many cases not all the variables at play are observable.
These hidden factors subtly change the observable data-generating distribution of the environment, 
causing inevitably a performance decay during the deployment. 
Indeed, when considering real-world applications, it is easy to assume that the environments where the model is deployed could see new data in the future with a different data distribution than that observed in training. 
Clearly, collecting another dataset and re-training the model is not a definitive solution as long as the previously acquired knowledge is not preserved. In fact, additional training on new data (Fine-Tuning) causes the effect known as Catastrophic Forgetting, where the model's performance on previous data deteriorates \cite{de2021continual, mccloskey1989catastrophic}. Continual Learning overcomes this limitation by enabling models to learn from new data without forgetting previously learned information.

\subsubsection{Continual Learning scenarios}

CL typically aims to learn a sequence of tasks, with each task representing a complete ML problem, usually in the form of classification. However, the wide variety of what constitutes a valid task sequence often makes it challenging to compare results across different studies. To solve this issue, many works organized similar experimental setups into the so-called CL Scenarios. Essentially they are a definition of what aspect of the problem is allowed to change between the tasks, what input information is available and what are the expected outputs. Although in literature one can find many Scenarios that derive from the specific needs of the authors that proposed them, we can safely say that the vast majority of published works use one of the following: \textit{Task-Incremental Learning} (TIL)~\cite{vandevan2019}, involves training a model on a sequence of general tasks but relies on the availability at deployment time of the task identifier to let the model know which of the learned tasks it must use as reference to make the inference. When such information is not available one can rely on either \textit{Class-Incremental Learning} (CIL)~\cite{vandevan2019} or \textit{Domain-Incremental Learning} (DIL)~\cite{vandevan2019}. The difference between them is very articulated in their original paper. Still, operationally the first only allows for subsequent tasks with a disjoint set of classes. In contrast, the second requires tasks with the same set of classes but allows a domain shift. Lastly, another important scenario to model more complex environments is \textit{New Instances \& Classes} (NIC)~\cite{lomonaco2017core50}, which allows both to update the model knowledge with the introduction of additional classes and domain shifts to expand the model knowledge of previously learned concepts (without the availability of the task identifier).

\subsubsection{Continual Learning approaches}
In the CL literature, the methods can be grouped into three big families of approaches known as rehearsal-based, regularization-based, and architecture-based.
\textbf{Rehearsal-based techniques} assume storing and reusing past data samples during training. 
While various approaches exist in the following category, one of the most renowned methods is  Experience Replay (ER) \cite{rolnick2019experience}, also known as Replay. 
With this method, samples from the Replay memory (which contains a small portion of the previously seen samples) are combined with the data of the new task. 
In this way, the model can retain the accumulated knowledge by periodically revisiting previous data while learning new information. 
This method has demonstrated remarkable performance in multi-class classification problems, consistently demonstrating its efficacy in all three CL scenarios (DIL, CIL, TIL).
\\
\textbf{Regularization-based approaches} consider additional constraints or penalties during training to maintain the memory of old tasks. 
Two well-known methods belonging to this category are Elastic Weight Consolidation (EWC) \cite{kirkpatrick2017overcoming} and Learning without Forgetting (LwF) \cite{li2017learning}.
The rationale behind EWC is to assign higher penalties to changes in parameters that are important for previously learned tasks, thus protecting them from significant updates during the training of the next tasks. 
\\
LwF is based on the concept of \textit{Knowledge Distillation}, introduced by Hinton et al.~\cite{hinton2015distilling}. The loss function incorporates both the classification error for the current task and a distillation loss, which uses the previous model’s output as soft labels. This approach encourages the model to produce consistent outputs across both old and new tasks, helping it retain prior knowledge while learning new information. 
\\
\textbf{Architecture-based approaches} modify the original model's architecture to preserve existing knowledge. The specific methods used in these approaches vary widely \cite{rusu2016progressive, fernando2017pathnet, mallya2018packnet}. However, these methods generally have two main limitations: they require memory that increases with the number of tasks, and they often necessitate the task ID during inference. These factors render them unsuitable for our objectives.

Moreover, in the CL literature, some methods propose combining different types of approaches to obtain better and more robust performance \cite{soutifcormerais2023comprehensiveempiricalevaluationonline}.
For example, the DER approach \cite{buzzega2020darkexperiencegeneralcontinual} combines the idea of Replay with the idea of Lwf about using the logits of previous models.

\section{Methodology}
\label{sec:methodology}
In Sec. \ref{subsec:problem_definition} we define the methodological framework designed to integrate Continual Learning into behavior-based driver identification systems. 
Then in Sec. \ref{subsec:scenarios_description}, three progressively realistic scenarios are introduced, where to evaluate the CL methods in the context of driver identification. 
In Sec. \ref{subsec:methods_used} we present the CL methods used in our experiments to assess the suitability of CL for addressing the driver identification problem on a data stream. 
Finally, in Sec. \ref{sec:pred_smoothing} our proposed approaches, SmooDER and SmooER, are described.

\subsection{Problem Definition}
\label{subsec:problem_definition}
As mentioned, Driver Identification is an important problem with numerous real-world applications. However, despite existing methods achieving impressive results in previous studies, they struggle to efficiently acquire new knowledge over time, as retraining a new model from scratch using all the seen data is resource-intensive and impractical. This limitation raises concerns about the feasibility of implementing them in real-world scenarios due to their scalability issues.

In a generic scenario, a model is trained on a sequence of $T$ tasks. Each task $t$ is associated with a dataset $D_t = (X_t, Y_t)$, where $X_t = [x_{1t}, \dots, x_{nt}]$ represents the driving samples for task $t$. Each sample $x_{it} \in \mathbb{R}^{W \times F}$ is a matrix recording the values of $F$ features of interest across $W$ consecutive time steps during the drive. The corresponding labels are stored in $Y_t = [y_{1t}, \dots, y_{nt}]$ with $y_{it} \in C_t$, where $C_t$ is the set of classes for task $t$.

The objective of the model at task $t$, $f_{\theta_t}$, is to learn a mapping:
\begin{equation}
f_{\theta_t}: x \rightarrow y \quad \text{such that} \quad x \in \mathbb{R}^{W \times F}, \quad y \in \bigcup_{i=1}^t C_i
\end{equation}
where $\bigcup_{i=1}^t C_i$ is the set of all classes encountered up to and including task $t$.
Therefore, after training the model for the task $t$, it is expected that it will perform well on all tasks, old and new.

\subsection{Scenarios}
\label{subsec:scenarios_description}
We propose and study the following three CL scenarios in the context of Driver Identification.
For each scenario, the goal is to accurately identify new drivers while maintaining the ability to correctly classify previously seen drivers.
However, each scenario has an increasing level of complexity and realism. 

\subsubsection{Scenario 1 - Two New Drivers}
\label{subsec:scenario1_description}
\begin{figure}[thbp]
  \centering
    \includegraphics[width=\textwidth]{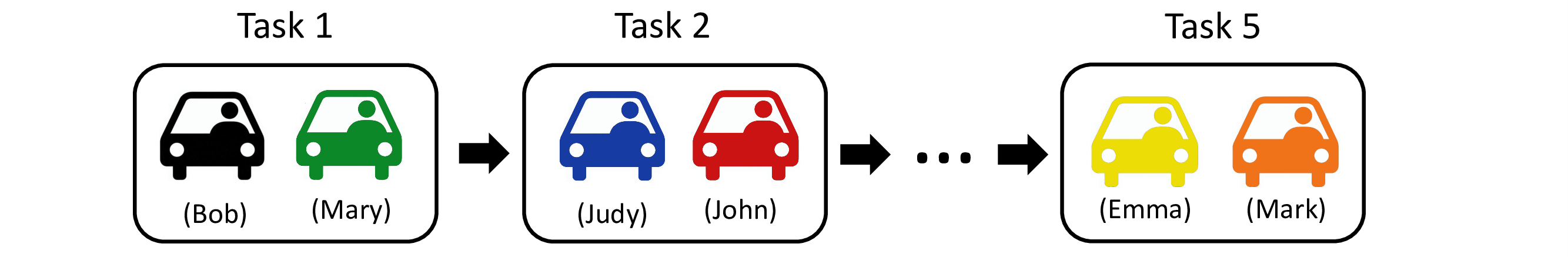}
    \caption{Task sequence representation for Scenario 1. Given ten drivers, each task contains information about two new drivers, for a total of five tasks.}
    \label{fig:CL_scenario1}
\end{figure}

    The first considered scenario is the simplest one. The stream is created similarly to those in classic CL benchmarks like Split-MNIST and Split-CIFAR10, where the dataset is split into 5 tasks, each composed of data of only two classes \cite{vandevan2019}.
    This scenario replicates the typical conditions under which CL techniques are usually proposed.
    Therefore, we split the OCSLab dataset into five tasks, each composed of two drivers' data (see Fig. \ref{fig:CL_scenario1}).    
    Since for each task, two new classes are added, it fits the CIL setting (described in Section \ref{subsec:continual_learning}).

\subsubsection{Scenario 2 - One New Driver}
\begin{figure}[thbp]
  \centering
    \includegraphics[width=\textwidth]{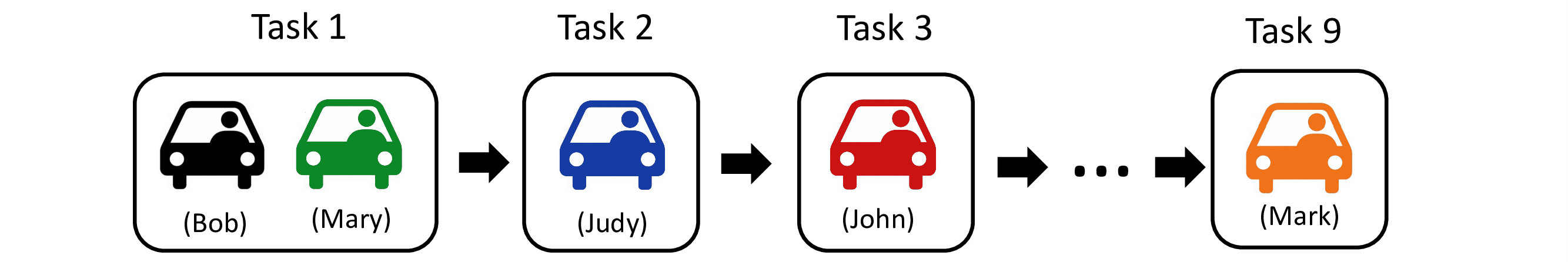}
    \caption{Task sequence representation for Scenario 2. A single driver is added to the system each time (except for task 1).}
    \label{fig:CL_scenario2}
\end{figure}

        In practice, there is no particular reason to add authorized drivers in pairs like in Scenario 1. However, in cases like a family car, it’s reasonable to expect that the owner might need to authorize an additional driver, such as when a child obtains their driver’s license. Therefore, for Scenario 2 we have a sequence of classes as follows:  2,1,1,1,1,1,1,1, for a total of 9 tasks (see Fig. \ref{fig:CL_scenario2}).
        The first task is composed of two classes since evaluating accuracy on a single class at the start would be pointless. In future tasks, accuracy is computed on all classes (drivers) seen so far.
        Compared to Scenario 1, here more tasks are required to cover the same number of classes (9 instead of 5). As a result, more updates are needed to learn the same number of drivers, which can potentially increase forgetting due to the longer task sequence. Additionally, introducing one class at a time makes it more challenging for the model to learn disjoint representations for each driver, as they are learned separately. Since a new class is added with each task, this scenario fits the CIL setting, as described in Section \ref{subsec:continual_learning}.
        
\subsubsection{Scenario 3 - Two New Sessions}
\label{subsec:scenario3_description}
    In this section, we outline a scenario that we believe is the most suitable for real-world implementation of a driver identification solution.

    A critical aspect of adding a new user to the system is the process of collecting training data, as the behavior-based driver identification system must be temporarily disabled, relying only on conventional identification to allow the new user to drive. Traditional data collection methods, such as those used in the OCSLab dataset, typically involve planning lengthy driving routes for each driver, with multiple sessions needed to accurately capture individual driving styles. This process is demanding for users, often requiring hours of driving, and lacks flexibility. The end result is a sequence of driving sessions from various drivers, often with repeated sessions for consistency.

    To alleviate this burden, we propose a system in which the conventional and behavior-based components complement each other. When introducing a new user, a startup phase defined by a super-user relies exclusively on conventional identification. After this initial phase, the system can automatically store a buffer of data from recent driving sessions. If the predictions from the conventional and behavior-based identification converge, the collected data can be utilized for additional training to adapt to domain shifts. Conversely, discrepancies can trigger an error notification. This setup enables continuous online training. The most efficient time to initiate training using the buffered data is when the car is parked for an extended period, avoiding unnecessary resource usage while driving. This approach can be naturally represented using the OCSLab dataset by segmenting the sequence of driving sessions into tasks. Only tasks involving the introduction of a new driver require access via traditional identification methods, enhancing security in the others.

    To formalize our scenario, two aspects are particularly important: the frequency of updates and the consistency of the task structure. Increasing the update frequency enhances the security of the system but also places greater demands on computational resources. Additionally, ensuring a fixed number of sessions per task is crucial for ensuring comparability of results. 
    
\begin{figure}[thbp]
  \centering
    \includegraphics[width=1.1\textwidth]{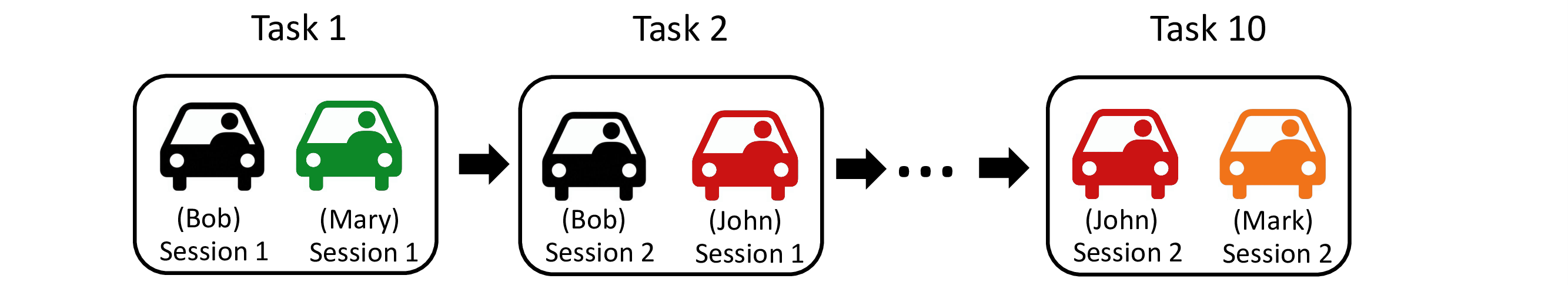}
    \caption{Task sequence representation for Scenario 3. The data is divided into drive sessions (there are two of them for each driver) and each task contains the data of two sessions. So the next task can either introduce new drivers or expand the knowledge of previously seen drivers.
}
    \label{fig:CL_scenario3}
\end{figure}

    We propose using two driving sessions per task, as depicted in Figure \ref{fig:CL_scenario3}, since this configuration covers the vast majority of cases. In real-world scenarios, this could represent either a short round trip by a single driver or a longer journey involving a driver change or a stop. Given that the dataset contains 20 driving sessions (two per driver), this setup corresponds to a total of 10 tasks.

    Unlike the previous scenarios, which fall under the CIL setting, this scenario aligns with the NIC setting, as described in Section \ref{subsec:continual_learning}. Here, tasks may include new driving sessions from previously seen drivers or data from entirely new drivers, making it a dynamic and realistic representation of the driver identification problem.

\subsection{Methods}
\label{subsec:methods_used}
Our goal is to validate, across three distinct scenarios, whether CL techniques can effectively address the Driver Identification problem. 
Therefore, we select three baselines, Fine-Tuning as the lower bound, and Joint Training and Cumulative as the upper bounds for model performance. 
Additionally, we test several well-known CL techniques such as ER, EWC, LwF, and DER.
\label{sec:cl_methods_theory}
\begin{itemize}
    \item \textbf{Joint Training}: The model is trained on all classes simultaneously, representing the traditional static training approach. This serves as an upper baseline for CL strategies, as it is not affected by catastrophic forgetting.
    \item \textbf{Cumulative}: This represents an additional upper bound of the CL techniques. The model is trained sequentially on the stream of tasks. However, contrary to Fine-Tuning, the model at each task is trained with all data seen so far (including all samples of the previous tasks). 
    The cumulative approach represents the current solution for adding additional users; however, it presents scalability challenges when expanding to a larger number of users.
    \item \textbf{Fine-Tuning}: The model is presented sequentially with data from the current task. No particular technique maintains the model’s knowledge, so it is considered a lower bound for the CL approaches.
    \item \textbf{ER}: Experience Replay creates a memory to store samples belonging to previously seen tasks \cite{rolnick2019experience}. We consider a fixed memory size over time and the same size dedicated to each task. When a task needs to be incorporated into the replay memory, some samples from the memory are randomly removed. 
    \item \textbf{EWC}: Elastic Weight Consolidation relies on the Fisher information matrix to identify which neural network parameters are important for the tasks the model has already learned \cite{kirkpatrick2017overcoming}. The Fisher information matrix measures the sensitivity of the loss function to changes in the model parameters. The importance is used to make the important weights more rigid.
    \item \textbf{LwF}: Learning without Forgetting is the most well-known technique among the distillation-based approaches \cite{li2017learning}. In our implementation, at each training batch, a distillation loss is computed as the Mean Squared Error (MSE) between the logits of the teacher and the student. The distillation loss is then added to the model loss for the task.
    \item \textbf{DER}: Dark Experience Replay \cite{buzzega2020darkexperiencegeneralcontinual} is similar to ER as it keeps a small sample from the previous tasks to refresh the old knowledge when learning from new data, but instead of storing the ground truth labels, it stores the so-called "dark knowledge" generated by a previous version of the model (its output logits). The original paper proposes another variant of DER termed DER++, which stores both ground truth and the model output logits for each saved sample.
\end{itemize}

\subsection{Prediction Smoothing}
\label{sec:pred_smoothing}
With this work, we introduce two new continual learning methods derived from DER++ and ER, specifically Smoothen Dark Experience Replay (SmooDER) and Smoothen Experience Replay (SmooER). These methods are designed to take advantage of a key feature of the scenarios we consider, which is the natural continuity of the driver in real-world applications where it is impractical to change the driver every minute.

We needed a mechanism that could adaptively correct occasional classification errors due to forgetting without causing significant delays during driver changes. Given that the labels are expected to change infrequently, smoothing the logits enhances robustness, as multiple consecutive incorrect predictions are needed to significantly alter the integrated result. Consequently, we employed a causal rolling-window average of the model's output logits. 
Formally, let $W$ be the window size, $k$ the number of classes currently present, and $f_\theta$ be the current model. 
During inference, given the i-th sample, the model calculates its output $z_i = f_\theta(x_i)$, where $z_i = [z_{i1}, z_{i2}, \cdots , z_{ik}]$ represents the logit for each class, or in other words the confidence for the $i-th$ sample to be a member of each of the classes.
We then apply the causal rolling-window average of Equation \ref{eq:smoothing} to the outputs obtaining $z_i = [ \widetilde{z}_{i1}, \widetilde{z}_{i2}, \cdots , \widetilde{z}_{ik}]$. Finally, the final output will be the class with the greatest confidence as in Equation~\ref{eq:argmax}.

\begin{equation}
    \widetilde{z}_{ij} = \frac{1}{W} \sum_{r=i-W+1}^{i}z_{rj}
    \label{eq:smoothing}
\end{equation}
\begin{equation}
    \hat{y}_i = \argmax_{j \in [1 \cdots k]} \sigma( \widetilde{z}_{ij} )
    \label{eq:argmax}
\end{equation}
where  $\sigma(\cdot)$ represents the sigmoid activation function.
The trade-off is governed by the length of the smoothing window. A larger window is advantageous for noisy predictors, as it can suppress more misclassifications. Conversely, a smaller window suits a robust classifier, reducing errors during driver changes, as fewer new-driver predictions are needed to alter the output. For this study, we selected a window length of six, which means a maximum delay of 36 seconds (smoothing window size multiplied by the data windowing stride defined in \ref{sec:spec_ocslab}).

\section{Experimental setting}
\label{sec:experimental_setting}
We present the experimental setting to evaluate the performance of CL methods in the context of behavior-based driver identification. 
Specifically, Sec. \ref{sec:spec_ocslab} details the decisions and preprocessing strategies applied to the OCSLab dataset.
Then Sec. \ref{subsec:exp_setting_models} provides the details about the deep learning model considered during the training and Sec. \ref{subsec_exp_setting_cl_approaches} provides the hyperparameters used for the tested CL methods.

\subsection{Specifics on the OCSLab dataset}
\label{sec:spec_ocslab}
The dataset used in our tests comprises data from an experiment involving ten drivers (labeled from "A" to "J"), each of them completed two round trips under varied road conditions driving the same car, for a total of about 23 hours of driving. The dataset contains 94401 items recorded every second and each record contains 51 sensors' measurements (you can refer to the original paper \cite{ocslab} for more detail).

Our main objective is to test the applicability of CL techniques in the Driver Identification problem. Since CL is generally model-agnostic, we decided to build our system over one of the most cited works in this field \cite{like_them}. 

To preprocess our data, we removed the features with zero variance in the dataset and then used the mean and standard deviation normalization. 
This technique standardizes the features of a dataset, transforming each feature \( x \) using the formula $x' = \frac{x - \mu}{\sigma}$, where \( \mu \) is the mean of the feature, and \( \sigma \) is the standard deviation. This process centers the data around zero and scales it to have a unit variance, improving the performance and convergence of many machine learning algorithms.

The next crucial part in preprocessing data for a sequence classification task is to choose suitable parameters for the windowing of our data. 
\begin{itemize}
    \item The \textbf{window length} is how many consecutive samples a training instance is made of and consequently defines the complexity of the task to be learned as well as the richness in the causal relationship that the model can exploit.
    \item The \textbf{stride of the window} refers to the displacement between consecutive windows in the data, allowing for potential overlap. This parameter is crucial as it determines how often each data point appears in the final dataset. A smaller stride means more frequent appearances of the same point in different contexts, enhancing model generalization. However, it also increases the dataset size, which can extend training time. Additionally, a larger stride reduces model responsiveness during deployment, as it requires waiting for a new set of measurements equal to the stride size to generate a valid instance.
\end{itemize}
For these parameters, we used the configuration from the reference paper with the shortest window length and stride to create a less complex but more responsive model. Specifically, we set the window length to 60 and the stride to 6.

In our experiment, we utilized two different setups:
\begin{itemize}
    \item For the comparison between the CL algorithms and baselines, we strictly followed the reference paper's methodology: a 70\%-30\% split for training and testing samples and a total of 300 epochs per experience. This choice is based on the reference results showing that every model had converged by the 300th epoch.

    \item For hyperparameters tuning, we sampled half of the test data as validation for early stopping. This was crucial to get the results in a reasonable amount of time. To ensure our results were comparable to each other we used a single Nvidia Titan V GPU, while the time to train a model was considerably longer than a standard setup since the data was split into tasks and sometimes reused when memory was involved in the CL algorithm. With this setup, most training sessions concluded around the 50th or 100th epoch, while still giving reliable indicators for the optimal parameters.
\end{itemize}

Due to varying data points among drivers in the dataset, the number of windows per class was inconsistent (see Table \ref{tab:count_data}), which could affect results when changing class order in tasks. To reduce randomness from class ordering and sampling, instead of 5-fold cross-validation, we ran each experiment with 4 different seeds and 4 class order permutations.

\begin{table}[tbh]
\caption{Count of windows in training and test set for each driver's drive session.}
\label{tab:count_data}
\centering
\begin{tabular}{|c|c|c|c|}
\hline

\textbf{Driver ID} 
& \textbf{Session} 
& \textbf{Train Windows (70\%)}
& \textbf{Test Windows (30\%)} \\ 

\hline

0 & 1 & 367 & 156 \\
\hline
0 & 2 & 466 & 199 \\
\hline
1 & 1 & 773 & 330 \\
\hline
1 & 2 & 717 & 306 \\
\hline
2 & 1 & 455 & 194 \\
\hline
2 & 2 & 409 & 174 \\
\hline
3 & 1 & 701 & 300 \\
\hline
3 & 2 & 831 & 356 \\
\hline
4 & 1 & 509 & 217 \\
\hline
4 & 2 & 463 & 198 \\
\hline
5 & 1 & 677 & 298 \\
\hline
5 & 2 & 596 & 255 \\
\hline
6 & 1 & 404 & 173 \\
\hline
6 & 2 & 458 & 195 \\
\hline
7 & 1 & 567 & 242  \\
\hline
7 & 2 & 574 & 245 \\
\hline
8 & 1 & 427 & 182 \\
\hline
8 & 2 & 472 & 201 \\
\hline
9 & 1 & 417 & 178 \\
\hline
9 & 2 & 610 & 261 \\
\hline

\end{tabular}
\end{table}

\subsection{Models}
\label{subsec:exp_setting_models}
We chose LSTM as our model because it consistently performed the best in the reference paper's result tables \cite{like_them}. According to their description, the network structure is a two-layer LSTM, with each layer having 128 hidden states and a dropout rate of 0.5.

LSTMs are sequence-to-sequence models that transform the input into a sequence of hidden states \cite{hochreiter1997long}. To create a classifier, a fully connected layer is applied to the last hidden state of the final layer with a sigmoid activation function. 
We trained our model with a batch size of 32 using Adam as an optimizer \cite{kingma2014adam}, following the reference paper, with a standard learning rate of 0.001.

\subsection{CL approaches}
\label{subsec_exp_setting_cl_approaches}
\label{subsec:hps}
Here, we describe the specific hyperparameters considered for each CL approach, their purpose, and the final value used in the method comparison. Unless otherwise noted, all methods use the learning rate, optimizer, and batch size values outlined in the previous section. 

\begin{itemize}    
    \item \textbf{ER}: This method needs two hyperparameters. The first is \textit{Memory Size}, which is also the most important as it is directly related to the resource constraints. For this parameter, we considered the optimal value \textbf{1000} as it lets us work with few resources, and at the same time, the loss in performance with respect to the Joint training baseline was not so drastic. The second hyperparameter is then \textit{Replay Ratio}, it essentially governs the mix ratios of instances from the training set and the memory in a batch of fixed size. Its impact is marginally on the performance and mainly on the time required for training, as the training set is diluted into more and more iterations when the ratio of memory instances increases. Here we chose to mix the two in equal parts, so its final value is \textbf{0.5}.
    
    \item \textbf{EWC \& LwF}: They both have a single hyperparameter $\lambda$ that is used as a weight to balance the influence of their regularization loss before adding it to the training loss. The best values that resulted from our experiments are \textbf{10000} for EWC and \textbf{5} for LwF.
    
    \item \textbf{DER++}: This method includes both rehearsal and regularization components. Similar to ER, the more important hyperparameter is the rehearsal memory size. Despite needing slightly more memory than ER to store additional logits, we chose to keep the optimal value at \textbf{1000}. For the parameters $\alpha$ and $\beta$, which balance the losses on memory logits and ground truth labels, the optimal values were \textbf{1} for both, indicating that the additional losses hold equal importance to the training loss.
\end{itemize}

\section{Results}
\label{sec:results}
In the following we present the results for each scenario, highlighting the performance of all tested CL methods and comparing existing approaches with our SmooER and SmooDER methods.

In Sec. \ref{subsec:scenario1}, we discuss results for the \textit{Two New Drivers} scenario, followed by the \textit{One New Driver} scenario in Sec. \ref{subsec:scenario2}, and the \textit{Two New Sessions} scenario in Sec. \ref{subsec:scenario3}. For each scenario, we provide two figures, with the left one comparing all the methods and baselines mentioned and the right one focusing on only the best non-baseline methods. A summary table also summarizes all the performances after learning the last task. We conclude with an overall analysis of the conducted experiments and results provided in Sec. \ref{subsec:overall_analysis}.

\begin{figure}[thbp!]
  \centering
  \begin{subfigure}[b]{0.49\textwidth}
    \includegraphics[width=\textwidth]{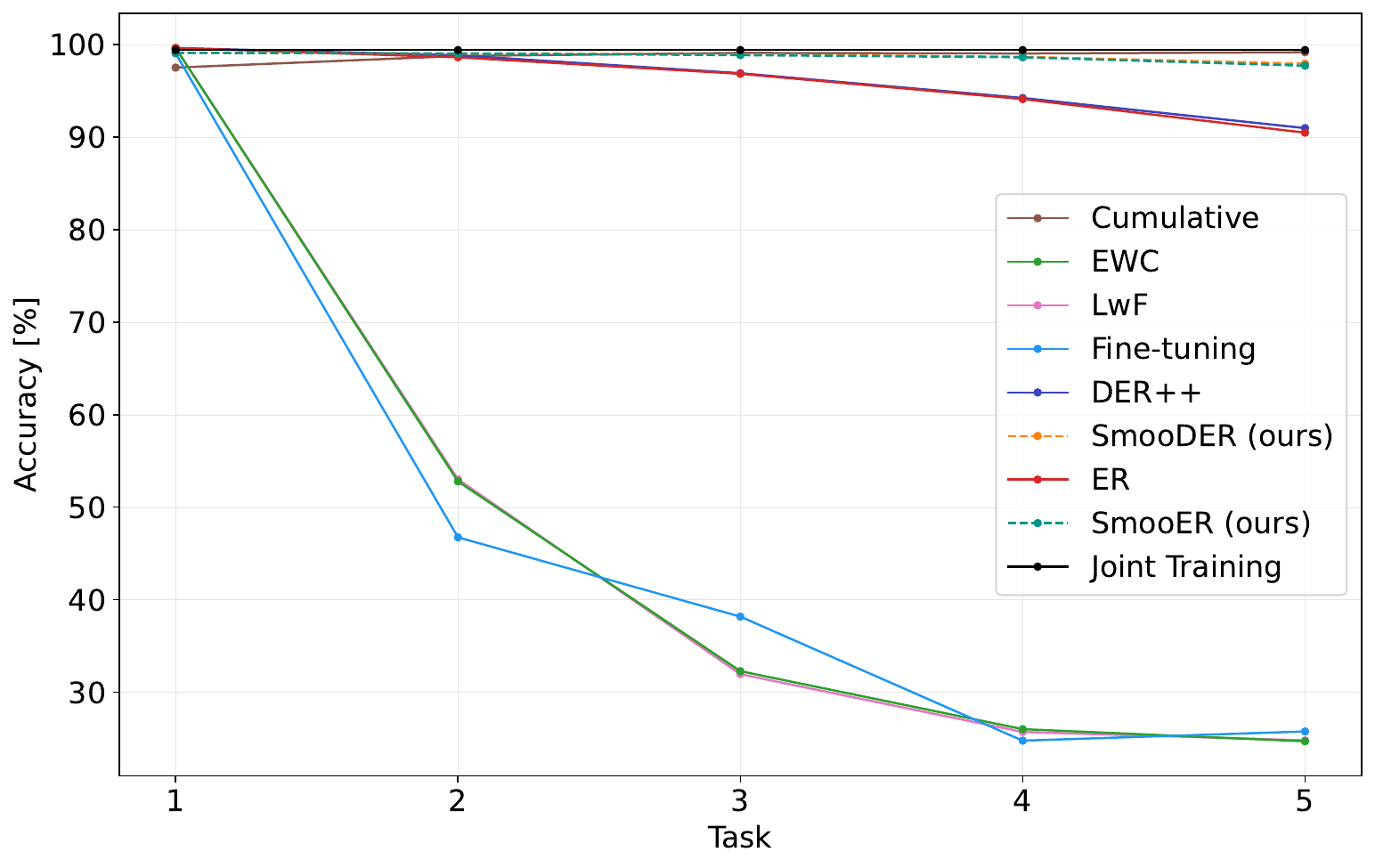}
    \caption{Test accuracy of each method and baseline.}
    \label{fig:res_acc_1}
  \end{subfigure}
  \hfill
  \begin{subfigure}[b]{0.49\textwidth}
    \includegraphics[width=\textwidth]{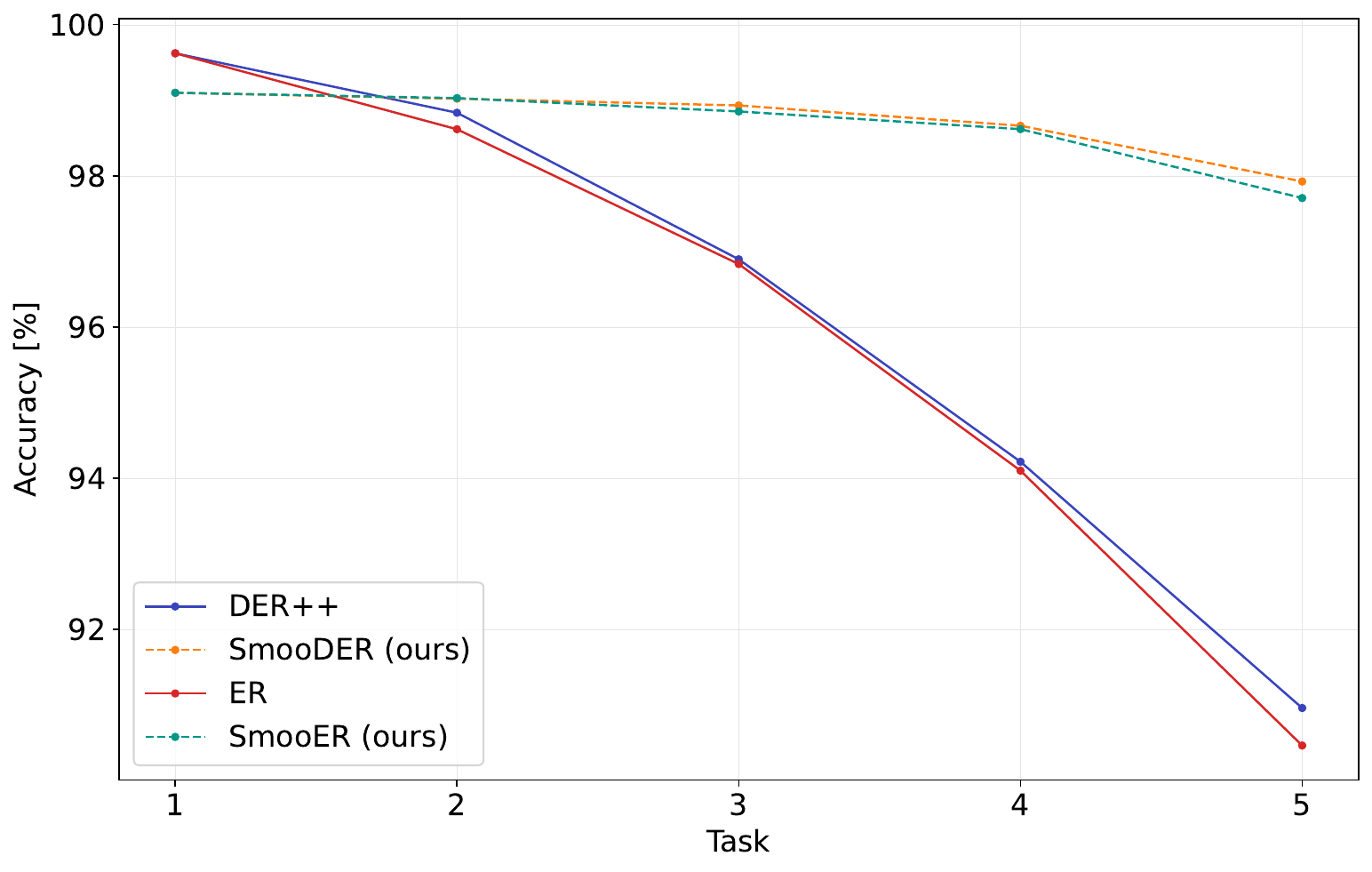}
    \caption{Focus on the best methods.}
    \label{fig:res_smo_1}
  \end{subfigure}
  \caption{Results for Scenario 1.}
  \label{fig:res_1}
\end{figure}

\subsection{Scenario 1 - Two New Drivers}
\label{subsec:scenario1}
This scenario has the least complexity among the three considered due to its shorter sequence length. As sequence length increases, the effects of catastrophic forgetting are expected to become more pronounced. As you can see in Table \ref{Tab:tab_results} this results in the final accuracy for this scenario to be the highest among the three. Corresponding to the classic Split-CIFAR10, its role is to link the CL literature with our work before adding the additional challenges in Scenarios 2 and 3 to simulate real-world applications more closely.

From the evolution of the accuracy along the sequence of tasks reported in Figure \ref{fig:res_acc_1}, we can distinguish three groups of methods: purely regularization-based methods like LwF and EWC have no effect in contrasting forgetting in this context as their evolution is very similar to Fine-Tuning. ER and DER++ instead demonstrate significantly better performance, with only about a 9\% drop in accuracy from the upper bound of 99.4\% set by Joint Training (Table \ref{Tab:tab_results}). Lastly, our proposed methods SmooDER and SmooER, further improve resistance to forgetting, achieving approximately 98\% reducing the gap of less than 2\%. In Figure \ref{fig:res_smo_1} you can see in more detail the advantage our methods have over ER and DER++, translating to an additional 7\% accuracy gain.

\begin{table}[tbh]
\caption{Performance for each scenario and for each CL technique. The best accuracy for each scenario is highlighted in bold.}
\label{Tab:tab_results}
\begin{adjustbox}{center}
\begin{tabular}{|c|cc|cc|cc|cc|}
\hline

& \multicolumn{2}{c|}{\textbf{Scenario 1}} 
& \multicolumn{2}{c|}{\textbf{Scenario 2}} 
& \multicolumn{2}{c|}{\textbf{Scenario 3}} 
& \multicolumn{2}{c|}{\textbf{Average of}} \\ 
& \multicolumn{2}{c|}{\textbf{Two New Drivers}} 
& \multicolumn{2}{c|}{\textbf{One New Driver}} 
& \multicolumn{2}{c|}{\textbf{Two New Sessions}} 
& \multicolumn{2}{c|}{\textbf{all Scenarios}} \\ 

\hline

& \multicolumn{1}{c|}{\textbf{ACC $\uparrow$}} & \textbf{gap $\downarrow$} 
& \multicolumn{1}{c|}{\textbf{ACC $\uparrow$}} & \textbf{gap $\downarrow$} 
& \multicolumn{1}{c|}{\textbf{ACC $\uparrow$}} & \textbf{gap $\downarrow$} 
& \multicolumn{1}{c|}{\textbf{ACC $\uparrow$}} & \textbf{gap $\downarrow$}  \\ 

\hline

\textbf{Cumulative}    
& \multicolumn{1}{c|}{99.16} & \multicolumn{1}{c|}{0.24} 
& \multicolumn{1}{c|}{98.96} & \multicolumn{1}{c|}{0.44} 
& \multicolumn{1}{c|}{97.43} & \multicolumn{1}{c|}{1.97} 
& \multicolumn{1}{c|}{98.52} & \multicolumn{1}{c|}{0.88} 
\\ 

\hline

\textbf{Joint Training} 
& \multicolumn{1}{c|}{99.40} & \multicolumn{1}{c|}{-}             
& \multicolumn{1}{c|}{99.40} & \multicolumn{1}{c|}{-}             
& \multicolumn{1}{c|}{99.40} & \multicolumn{1}{c|}{-}             
& \multicolumn{1}{c|}{99.40} & \multicolumn{1}{c|}{-}  \\ 

\hline

\textbf{Fine-tuning}    
& \multicolumn{1}{c|}{24.55} & \multicolumn{1}{c|}{74.85} 
& \multicolumn{1}{c|}{10.87} & \multicolumn{1}{c|}{88.53} 
& \multicolumn{1}{c|}{19.17} & \multicolumn{1}{c|}{80.23} 
& \multicolumn{1}{c|}{18.20} & \multicolumn{1}{c|}{81.20} 
\\ 

\hline\hline

\textbf{ER}         
& \multicolumn{1}{c|}{90.47} & \multicolumn{1}{c|}{8.93}  
& \multicolumn{1}{c|}{86.25} & \multicolumn{1}{c|}{13.15} 
& \multicolumn{1}{c|}{87.08} & \multicolumn{1}{c|}{12.32} 
& \multicolumn{1}{c|}{87.93} & \multicolumn{1}{c|}{11.47} 
\\ 

\hline

\textbf{EWC}            
& \multicolumn{1}{c|}{24.69} & \multicolumn{1}{c|}{74.71} 
& \multicolumn{1}{c|}{10.92} & \multicolumn{1}{c|}{88.48}
& \multicolumn{1}{c|}{19.10} & \multicolumn{1}{c|}{80.30} 
& \multicolumn{1}{c|}{18.24} & \multicolumn{1}{c|}{81.16} 
\\ 

\hline

\textbf{LwF}            
& \multicolumn{1}{c|}{24.78} & \multicolumn{1}{c|}{74.62} 
& \multicolumn{1}{c|}{10.88} & \multicolumn{1}{c|}{88.52} 
& \multicolumn{1}{c|}{20.01} & \multicolumn{1}{c|}{79.39} 
& \multicolumn{1}{c|}{18.56} & \multicolumn{1}{c|}{80.84} 
\\ 

\hline

\textbf{DER++}            
& \multicolumn{1}{c|}{90.96} & \multicolumn{1}{c|}{8.44} 
& \multicolumn{1}{c|}{87.04} & \multicolumn{1}{c|}{12.36} 
& \multicolumn{1}{c|}{88.87} & \multicolumn{1}{c|}{10.53} 
& \multicolumn{1}{c|}{88.96} & \multicolumn{1}{c|}{10.44} 
\\ 

\hline

\textbf{SmooDER (ours)}            
& \multicolumn{1}{c|}{\textbf{97.93}} & \multicolumn{1}{c|}{\textbf{1.47}}
& \multicolumn{1}{c|}{\textbf{95.47}} & \multicolumn{1}{c|}{\textbf{3.93}} 
& \multicolumn{1}{c|}{\textbf{98.22}} & \multicolumn{1}{c|}{\textbf{1.18}} 
& \multicolumn{1}{c|}{\textbf{97.21}} & \multicolumn{1}{c|}{\textbf{2.19}} 
\\ 
\hline

\textbf{SmooER (ours)}            
& \multicolumn{1}{c|}{97.71} & \multicolumn{1}{c|}{1.69} 
& \multicolumn{1}{c|}{95.26} & \multicolumn{1}{c|}{4.14} 
& \multicolumn{1}{c|}{97.45} & \multicolumn{1}{c|}{1.95} 
& \multicolumn{1}{c|}{96.81} & \multicolumn{1}{c|}{2.59} 
\\ 

\hline
\end{tabular}
\end{adjustbox}
\end{table}

\subsection{Scenario 2 - One New Driver}
\label{subsec:scenario2}
As described in Sec. \ref{subsec:scenarios_description}, considering a single new driver for each task to increase the realism resulted in a more challenging scenario as we are now dealing with twice the number of tasks. Table \ref{Tab:tab_results} shows that, as expected, the performance of all methods declined compared to Scenario 1, with the best-performing ones, ER and DER++, experiencing a reduction of around 4\% in accuracy. Notably, SmooDER and SmooER demonstrated even lower degradation, with a decrease of only about 2.5\%. Given the longer sequence and the complexity of learning each driver individually, this minor drop indicates that our methods have an acceptably low level of performance degradation as the number of tasks increases.

\begin{figure}[thbp!]
  \centering
  \begin{subfigure}[b]{0.49\textwidth}
    \includegraphics[width=\textwidth]{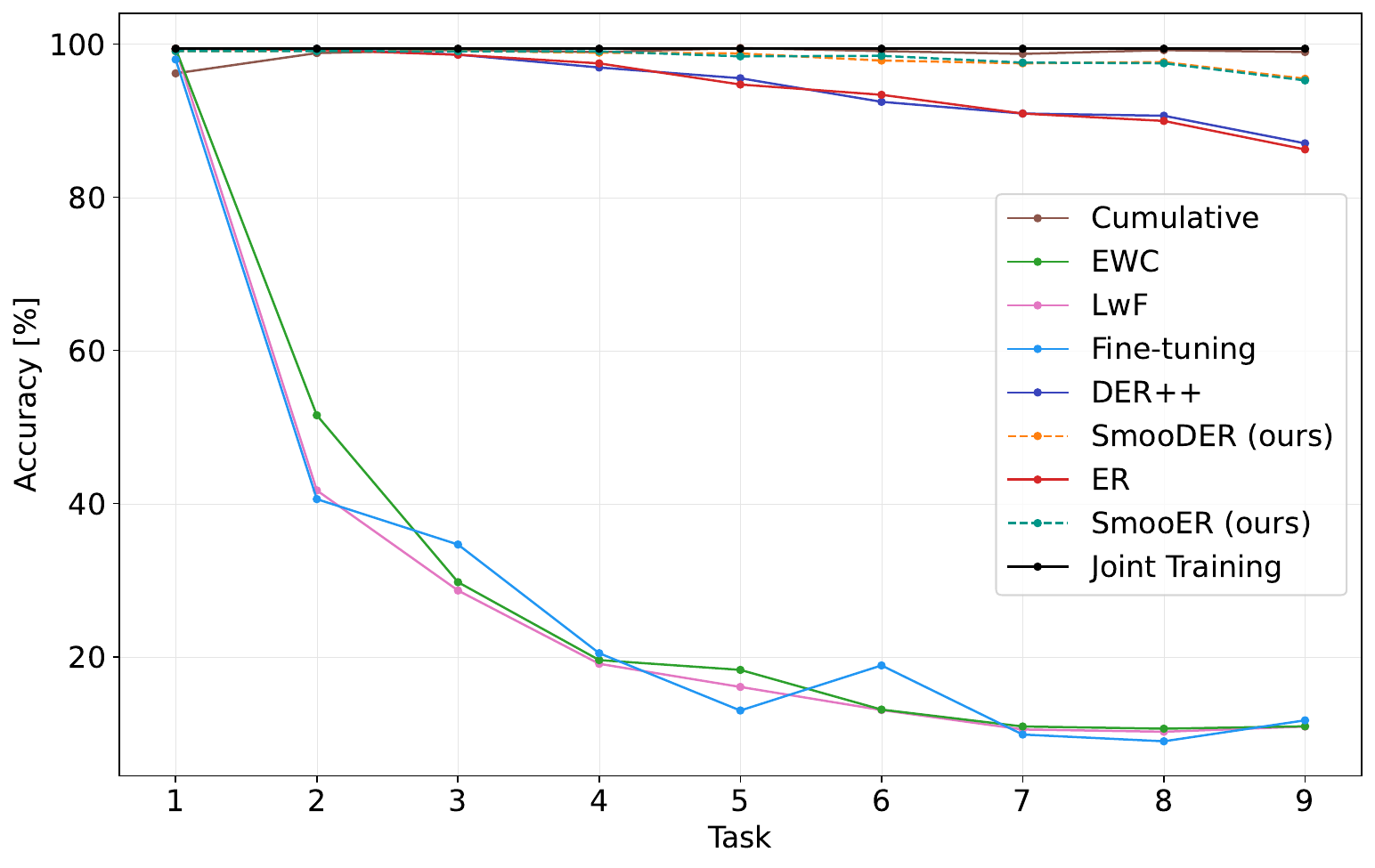}
    \caption{Test accuracy of each method and baseline.}
    \label{fig:res_acc_2}
  \end{subfigure}
  \hfill
  \begin{subfigure}[b]{0.49\textwidth}
    \includegraphics[width=\textwidth]{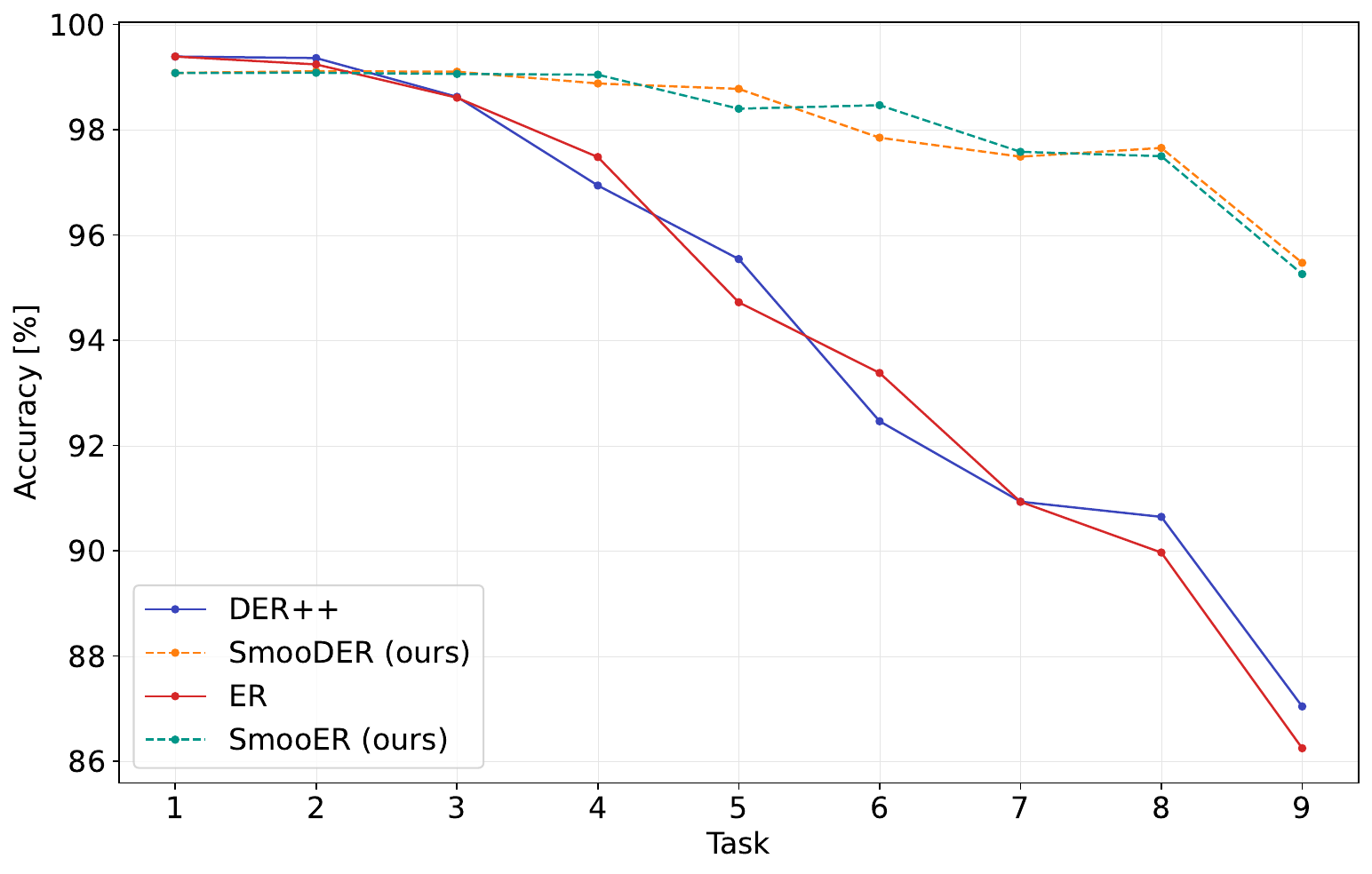}
    \caption{Focus on the best methods.}
    \label{fig:res_smo_2}
  \end{subfigure}
  \caption{Results for Scenario 2.}
  \label{fig:res_2}
\end{figure}

Figure \ref{fig:res_acc_2} presents an accuracy progression very similar to Scenario 1, with a notable observation: without effective countermeasures against forgetting, accuracy can drop as low as 11\%, as seen with LwF, EWC, and Fine-Tuning. In this more complex scenario, the accuracy gap between Joint Training and DER++ (similarly ER) has increased to about 13\%. While this gap might still be considered acceptable, it highlights the need for further improvement to mitigate forgetting and enhance performance. In this regard, our methods, SmooDER and SmooER, achieve a final accuracy of approximately 95\%, reducing the gap to Joint Training (99.4\%) to less than 5\%. This translates to an improvement of roughly 8\% over ER and DER++ which can be better appreciated from Figure \ref{fig:res_smo_2}, demonstrating the robustness of our approach in managing forgetting in more challenging scenarios.

\subsection{Scenario 3 - Two New Sessions}
\label{subsec:scenario3}
In Scenario 3, complexity is further increased by including partial data from two classes per task, resulting in a sequence of ten tasks. Here, each task may involve learning a new driver or expanding knowledge of an already encountered one. Despite this added complexity, the performance of various CL methods shows a slight improvement over Scenario 2. This observation provides two key insights: first, it suggests that the primary challenge still lies in the length of the task sequence. Second, the slight performance boost hints that learning multiple drivers per task enhances the model’s learning capabilities.

\begin{figure}[thbp!]
  \centering
  \begin{subfigure}[b]{0.49\textwidth}
    \includegraphics[width=\textwidth]{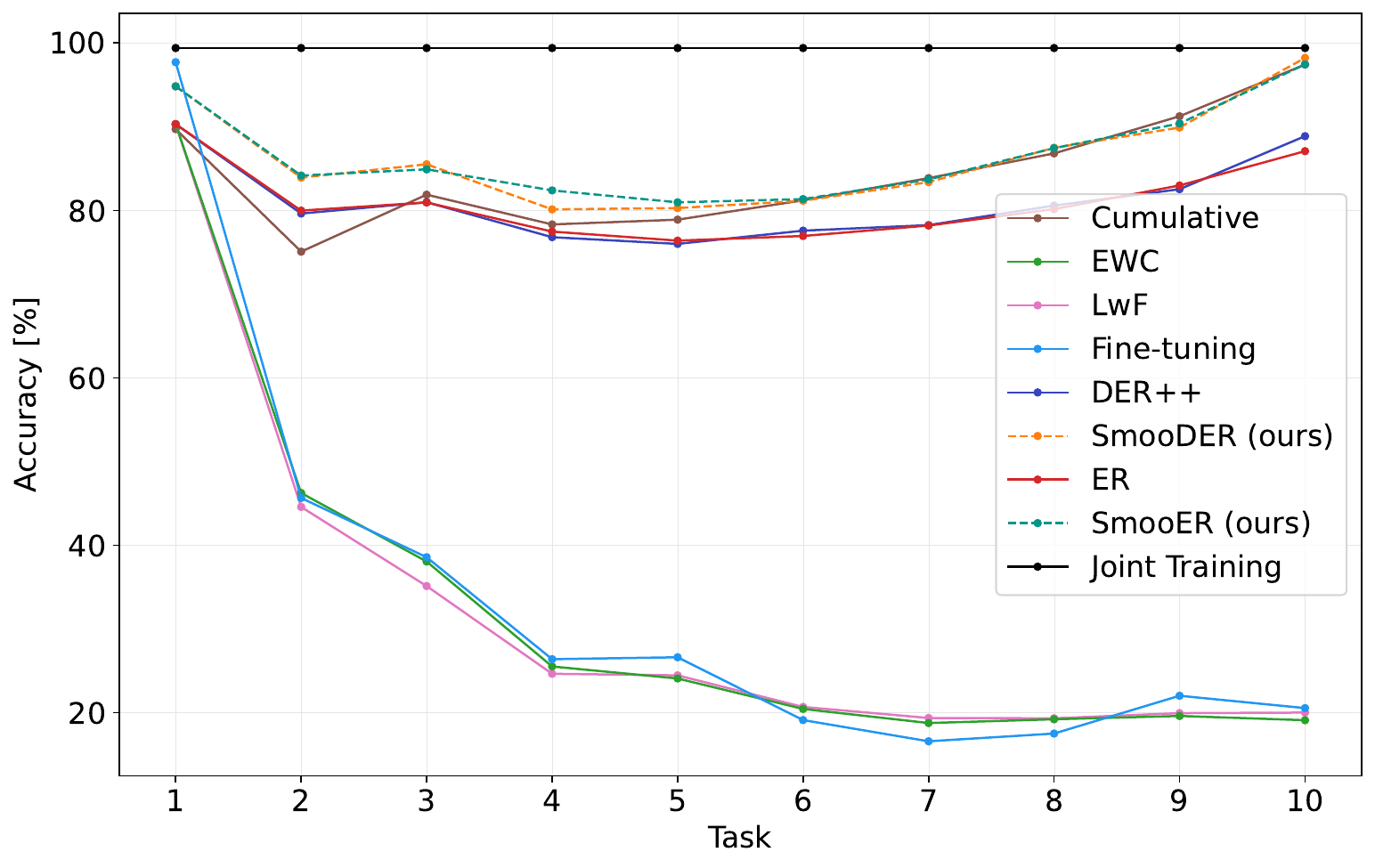}
    \caption{Test accuracy of each method and baseline.}
    \label{fig:res_acc_3}
  \end{subfigure}
  \hfill
  \begin{subfigure}[b]{0.49\textwidth}
    \includegraphics[width=\textwidth]{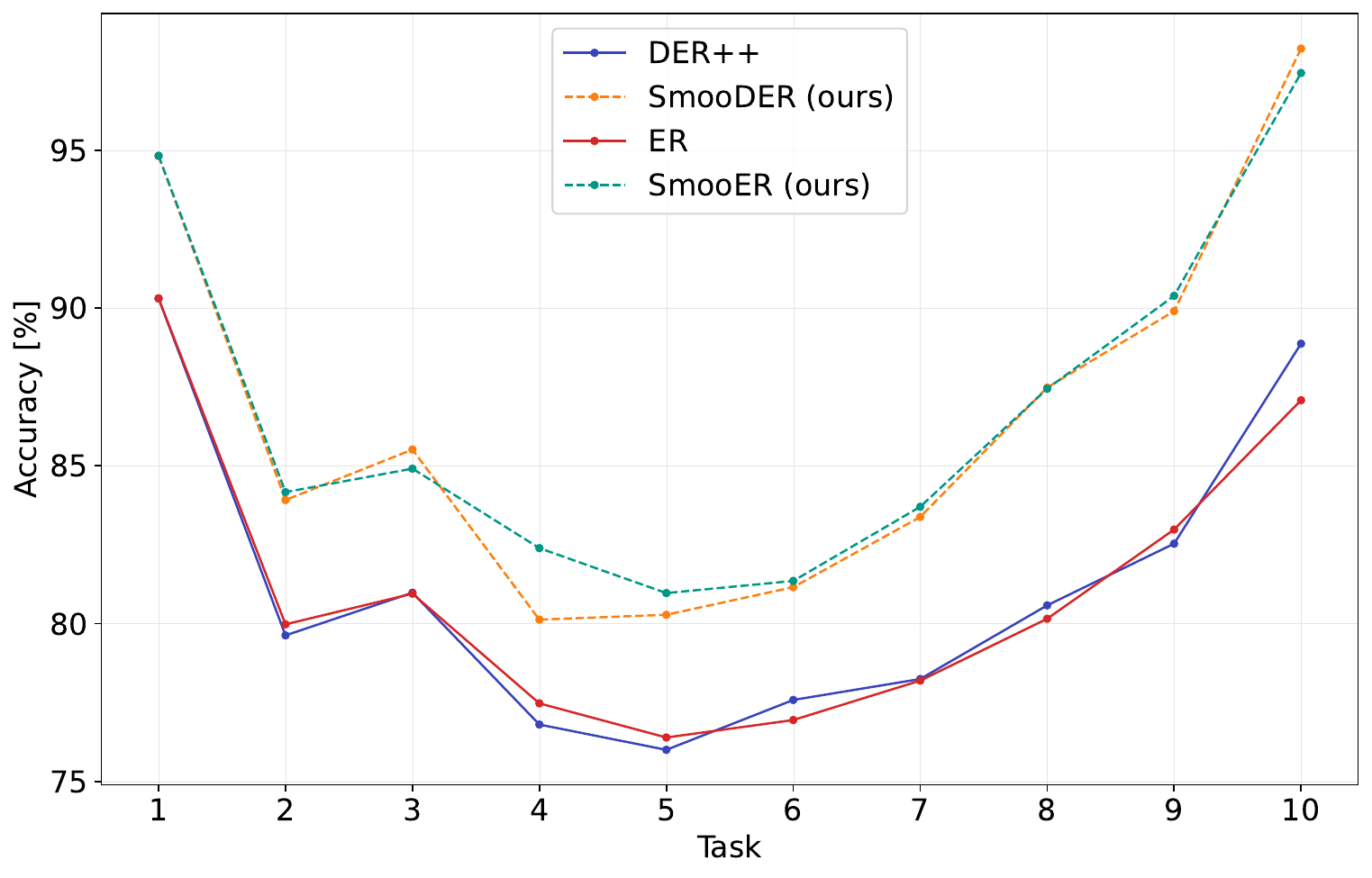}
    \caption{Focus on the best methods.}
    \label{fig:res_smo_3}
  \end{subfigure}
  \caption{Results for Scenario 3.}
  \label{fig:res_3}
\end{figure}

For this final scenario, Figure \ref{fig:res_acc_3} shows a different accuracy progression, further highlighting the effectiveness of methods capable of counteracting forgetting. Since the test set includes both driving sessions as soon as each driver is introduced (see Fig. \ref{fig:CL_driver}), accuracy tends to increase in later tasks as the model learns the last session for each driver. As a result, methods that effectively retain learned knowledge exhibit a \textit{U}-shaped accuracy trend, while those that struggle with forgetting like LwF and EWC do not.

Figure \ref{fig:res_smo_3} focuses specifically on methods that manage to retain previous knowledge. ER and DER++ achieve accuracy levels of 87\% and 89\%, respectively, maintaining a gap of around 10\% from Joint Training. Notably, our methods, SmooER and SmooDER, show an even better accuracy, reaching 97.45\% and 98.22\%, respectively, thus narrowing the gap to the upper bound to approximately 2\%. As a final consideration, the random inclusion of domain-shift elements—when the second driving session is introduced for each driver—likely contributes to the approximately 10\% performance improvement of methods like EWC and LwF compared to Scenario 2.

\subsection{Overall Analysis}
\label{subsec:overall_analysis}
In summary, SmooDER consistently outperforms all other methods across all three scenarios. The average results presented in Table \ref{Tab:tab_results} indicate that the DER++ and ER achieve average accuracies of 88.96\% and 87.93\%, respectively, making them preferable to LwF and EWC. 
Their scores of 18.24\% and 18.56\% are only slightly better than the Fine-Tuning baseline, which yields an average accuracy of 18.20\%. When compared to the Joint Training reference, where all data is trained at once, even the best-performing CL methods still show a 10\% accuracy gap. In contrast, SmooDER and SmooER significantly narrow this gap to just 2\%, demonstrating their effectiveness in achieving near-optimal accuracy.

\begin{figure}[thbp]
  \centering
    \includegraphics[width=0.75\textwidth]{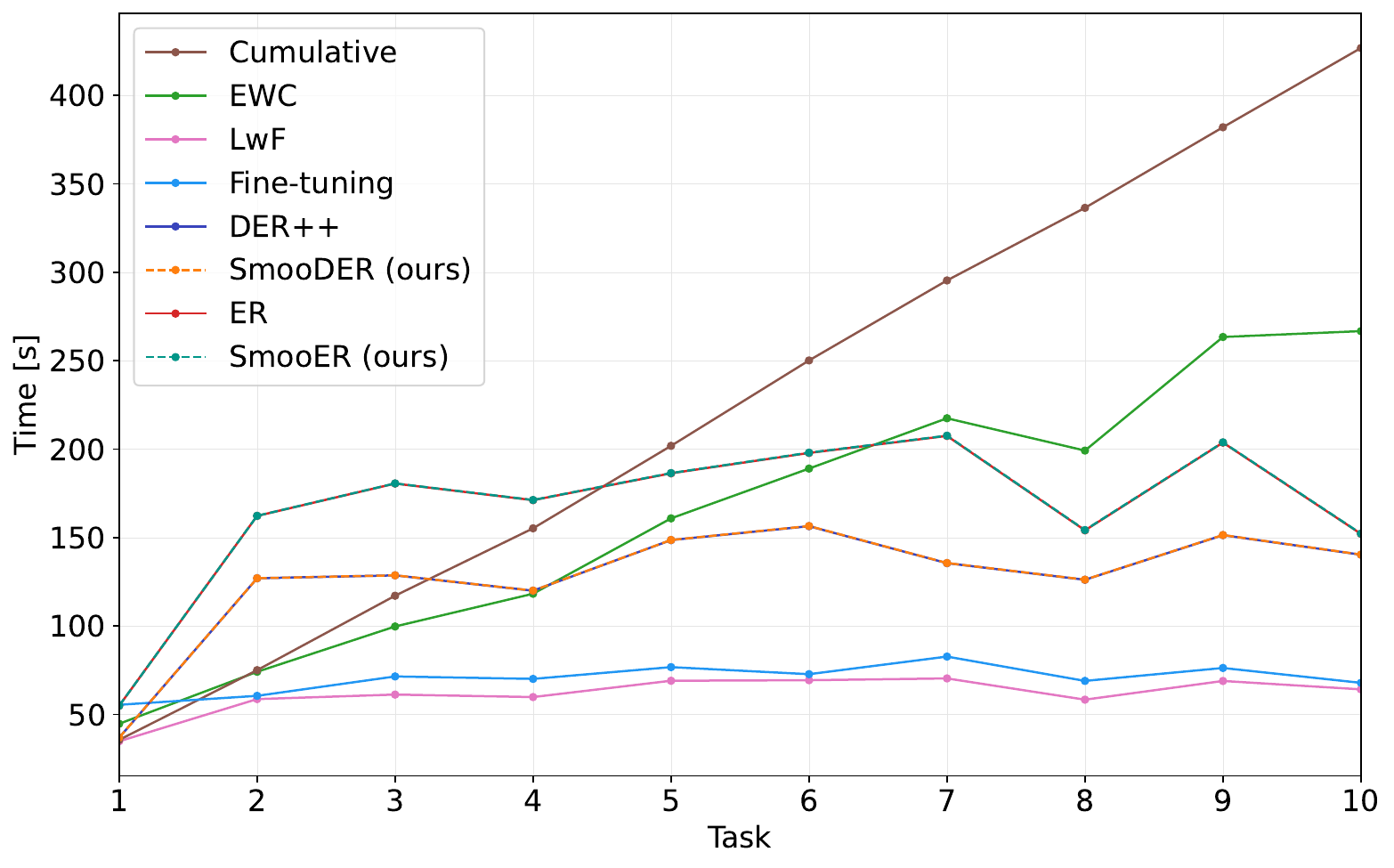}
    \caption{Training time per task for each method and baseline in Scenario 3. Cumulative training time increases linearly as more tasks are added since the datasets are concatenated. Fine-tuning and LwF have negligible or no computational overhead. EWC computational time grows since by implementation details each task contributes independently to the regularization. DER++, ER, SmooDER, and SmooER go to saturation when the memory is filled for the first time. The overhead of SmooDER over DER++ is negligible (same for SmooER over ER).}
    \label{fig:time_plot}
\end{figure}

Scalability in terms of both memory and time requirements is another key result, which we have not addressed in previous sections. Figure \ref{fig:time_plot} illustrates the training time (in seconds) for each task across all CL methods in Scenario 3. Among these, the Cumulative baseline, currently used for adding new users, achieves optimal accuracy by retraining from scratch with both old and new data. However, this approach leads to linearly increasing time requirements as tasks accumulate. In contrast, Fine-Tuning only processes new data, resulting in nearly constant training time across tasks. LwF, positioned between these approaches, also maintains constant time requirements, as it only computes a regularization term without significant extra computation. Meanwhile, EWC shows a growing time curve, as its implementation requires a separate weight matrix for each task. Alternative versions of EWC, which condense these matrices, are available to reduce training time. Memory-based methods like DER++ and ER offer a constant time curve, situated between Fine-Tuning and Cumulative, depending on memory size. It is important to note that CL methods do not take effect during the first task, so these considerations apply from task 2 onward. Lastly, SmooDER and SmooER require training and testing times similar to DER++ and ER respectively, as the additional computational overhead is minimal.

Focusing on the highest-performing methods, we see that Cumulative requires storing all previously encountered data, demanding 120 MB of memory to hold the complete dataset, which introduces significant practical constraints. By contrast, DER++ and ER (as well as their respective variants, SmooDER and SmooER, which are identical in memory requirements) only need to store a small subset of samples. In the results shown in Table \ref{Tab:tab_results}, we use 1,000 samples, occupying just 11 MB — specifically, less than $1000 \times (60 \times 46 \times 4B + 8B + 10 \times 4B) = 11,088,000$ bytes for DER++ and $1000 \times (60 \times 46 \times 4B + 8B) = 11,048,000$ bytes for ER. Including the network size of 0.89 MB, a total of only 12 MB of memory is needed. Moreover, for the deployment environment, just 6 MB of RAM is sufficient: 0.89 MB for the network, 3.93 MB for the backward pass during training, and 0.35 MB for the input batch. This efficient memory usage enables direct deployment on small devices, such as those commonly installed in vehicles~\cite{infineon_aurix_tc39xxx}.

\section{Conclusion}
\label{sec:conclusion}
In this work, we addressed the challenge of Driver Identification. We moved beyond the traditional approach, where all driver data is collected before model deployment, and training occurs offline. In real-world scenarios, such as family cars where additional relatives may start driving over time or car-sharing services that continually add new drivers as they sign up, this approach is inadequate. Continual learning allows the system to not only add new drivers as they come but also to incorporate new data from previously seen drivers. This approach enables a system that progressively adapts and enhances its performance as new data is collected.

Therefore, our work proved the feasibility of learning from a stream of new data and drivers by applying well-known CL techniques. 
Specifically, we evaluate the CL approaches on the commonly used OCSLab dataset.
Considering several scenarios of increasing realism and complexity, we tested and proved that the CL methods can learn effectively through a stream of new data and drivers.

Moreover, we considered the temporal correlation among the samples during the inference phase and proposed two novel methods, SmooDER and SmooER.
By testing our novel methods across all the scenarios, we demonstrate that SmooDER and SmooER significantly improve performance. 
Specifically, SmooDER achieves 97.21\% of accuracy compared to 88.96\% of DER++, while SmooER has 96.81\% compared to 87.93\% of ER. 
Our research successfully demonstrates the feasibility of learning new drivers from a stream directly on the car, reducing the gap with the static setting to only 2.19\% and 2.59\% for SmooDER and SmooER, respectively.
While our work improved the methodology of CL and proved its feasibility for the Driver Identification problem, several future potential research directions exist.
Though the smoothing is very effective, the proposed approach is simple, but more complex techniques can reduce even more effectively misclassification.
In this work, we considered a fixed smoothing window size. 
However, using an adaptive window size (or multiple window sizes simultaneously) could improve performance.

Eventually, developing systems resistant to spoofing or manipulation ensures the identified driver is indeed operating the vehicle. 
There is significant potential for future research into the effects of adversarial and data poisoning attacks on behavior-based driver identification systems that use continual learning. 
Adversarial attacks may cause misclassifications, while data poisoning can degrade model performance over time. 
Continual learning increases vulnerability by updating the model with new data, making it susceptible to gradual manipulation. 
Therefore, future works should focus on developing defenses that balance adaptability and robustness, particularly in the context of continual learning.

\bibliographystyle{unsrt}  
\bibliography{references}

\appendix

\end{document}